\documentclass{article} % DO NOT CHANGE THIS
\usepackage[preprint]{aaai25}  % DO NOT CHANGE THIS
\usepackage{times}  % DO NOT CHANGE THIS
\usepackage{helvet}  % DO NOT CHANGE THIS
\usepackage{courier}  % DO NOT CHANGE THIS
\usepackage[hyphens]{url}  % DO NOT CHANGE THIS
\usepackage{graphicx} % DO NOT CHANGE THIS
\usepackage{natbib}  % DO NOT CHANGE THIS AND DO NOT ADD ANY OPTIONS TO IT
\usepackage{caption} % DO NOT CHANGE THIS AND DO NOT ADD ANY OPTIONS TO IT
\usepackage{algorithm}
\usepackage{algpseudocode}

\usepackage{newfloat}
\usepackage{listings}
\DeclareCaptionStyle{ruled}{labelfont=normalfont,labelsep=colon,strut=off} % DO NOT CHANGE THIS
\floatstyle{ruled}
\newfloat{listing}{tb}{lst}{}
\floatname{listing}{Listing}
\usepackage{verbatim}

\title{Unveiling Population Heterogeneity in Health Risks Posed by Environmental Hazards Using Regression-Guided Neural Network}
\author{
    Jong Woo Nam\textsuperscript{\rm 1},
    Eun Young Choi\textsuperscript{\rm 2},
    Jennifer A. Ailshire\textsuperscript{\rm 2},
    Yao-Yi Chiang\textsuperscript{\rm 3}
}
\affiliations{
    \textsuperscript{\rm 1} Neuroscience Graduate Program, University of Southern California\\
    \textsuperscript{\rm 2} Leonard Davis School of Gerontology, University of Southern California\\
    \textsuperscript{\rm 3} Department of Computer Science and Engineering, University of Minnesota\\
    namj@usc.edu, choieuny@usc.edu, ailshire@usc.edu, yaoyi@umn.edu
}

\begin{document}
\frenchspacing

\maketitle
\begin{abstract}
Environmental hazards place certain individuals at disproportionately higher risks. As these hazards increasingly endanger human health, precise identification of the most vulnerable population subgroups is critical for public health. Moderated multiple regression (MMR) offers a straightforward method for investigating this by adding interaction terms between the exposure to a hazard and other population characteristics to a linear regression model. However, when the vulnerabilities are hidden within a cross-section of many characteristics, MMR is often limited in its capabilities to find any meaningful discoveries. Here, we introduce a hybrid method, named regression-guided neural networks (ReGNN), which utilizes artificial neural networks (ANNs) to non-linearly combine predictors, generating a latent representation that interacts with a focal predictor (i.e. variable measuring exposure to an environmental hazard). We showcase the use of ReGNN for investigating the population heterogeneity in the health effects of exposure to air pollution (PM2.5) on cognitive functioning scores. We demonstrate that population heterogeneity that would otherwise be hidden using traditional MMR can be found using ReGNN by comparing its results to the fit results of the traditional MMR models. In essence, ReGNN is a novel tool that enhances traditional regression models by effectively summarizing and quantifying an individual's susceptibility to health risks.
\end{abstract}

\section{Introduction}
In 2016, nearly 24\% of global mortality, corresponding to 14 million deaths per year, was attributable to environmental hazards \cite{pruss-ustunDiseasesDueUnhealthy2017}. These hazards include air pollution, severe weather events, and exposure to harmful chemicals. In the United States, the financial burden of health effects related to the environment was estimated at \$10.0 billion (2018 dollars) from ten climate-sensitive events in 2012 alone \cite{limayeEstimatingHealthRelatedCosts2019}. Mounting evidence indicates that environmental hazards are expected to escalate due to climate change, industrialization, and population growth. 

Although everyone can be exposed to environmental health threats, their impacts are not uniformly distributed. Vulnerable groups and communities, particularly those least equipped to anticipate, cope with, and recover from adverse impacts, face disproportionate and unequal risks \cite{ottoSocialVulnerabilityClimate2017}. Consequently, understanding and identifying population heterogeneity in environmental health risks is critical for developing targeted policies and interventions that address the unique vulnerabilities of different groups.

Moderated multiple regression (MMR) provides a simple means for investigating such disparities in health effects of environmental hazards \cite{saundersModeratorVariablesPrediction1956a}. 
MMR quantifies how an effect of a focal predictor variable \textit{x$_{f}$} on an outcome \textit{o} varies according to other independent variables \textit{m$_{k}$}, often referred to as moderators, by augmenting basic linear regression with their multiplied terms (${x_{f}*m_{k}}$).  These terms are referred to as interaction terms, as the two predictors \textit{interact} to moderate each other's effect sizes, predicting the outcome. 

This simple extension allows us to continue to utilize all statistical tools available for a linear regression model. For instance, a meaningful interaction between a moderator $m_k$ and the focal predictor $x_f$ is considered found when the regression coefficient $c^{int}_k$ for the interaction term $x_{f}m_k$ is statistically significant (p $<$ 0.05). The effect size of the moderation is quantified by the regression coefficient.

Many researchers have used MMR to discover the moderating effects of different population characteristics on the health effects of several environmental hazards. For example, MMR is used to find out that the mortality risks associated with extreme temperatures are significantly higher for women than for men \cite{son_temperature-related_2019}, and health risks associated with short-term exposure to particulate matter (PM) are substantially greater for older adults compared to younger ones \cite{bell_evidence_2013}. Despite its simplicity, MMR still is prevalent as a tool facilitating statistical quantification of uneven effects of health-related risk factors.

MMR, however, is not devoid of limitations. Firstly, due to the correlations between interaction terms \textit{x$_{f}$m$_{k}$} and their linear components (\textit{x$_{f}$} and \textit{m$_{k}$}), MMR often suffers from structural multicollinearity. Multicollinearity distorts the regression model by inflating its coefficients, potentially misleading interpretations of the data. While some studies point out that multicollinearity can be reduced by pre-processing the data and dismiss the necessity to control multicollinearity at all in MMR \cite{mcclellandMulticollinearityRedHerring2017}, we observe that the variance inflation factor (VIF) remains high in many cases even after necessary pre-processing steps are applied. This concerns scientists whether the model is estimating its coefficients correctly and also questions the findings using it.

Additionally, the effectiveness of MMR diminishes as more interaction terms are added, which reduces its statistical power \cite{aguinisStatisticalPowerModerated1995}. This necessitates judicious selection regarding the number and nature of interaction terms to include. Consequently, scientists are often driven by a strong scientific hypothesis when deciding which interaction terms to include in their regression model. Complexity increases if we want to consider higher-order-interactions (i.e. three-way-interaction, such as \textit{x$_{i}$x$_{j}$x$_{f}$}). 

Lastly, MMR proves inadequate in cases where the moderation effect is dispersed across multiple predictors rather than isolated to a single moderator. When multiple predictors are at interplay in moderating the effect of the focal predictor, it becomes impossible for a practitioner to guess which order and how many interaction terms to include in the model.

On the other hand, machine learning has recently emerged as a promising alternative to traditional methods. However, it is often criticized for lacking transparency and being hard to interpret. Studies like \cite{lampaIdentificationComplexInteractions2014, huangUseMachineLearning2023} use partial-dependence-based measures to tap into its complexity. However, because it lacks tools such as hypothesis tests to rule out spurious findings or means to decompose the influence of each predictor, practitioners in social science and health-related fields treat machine learning as a different breed, using it in contexts where prediction accuracy takes priority over statistical robustness. 

\begin{figure}
    \centering
     \includegraphics[width=\linewidth]{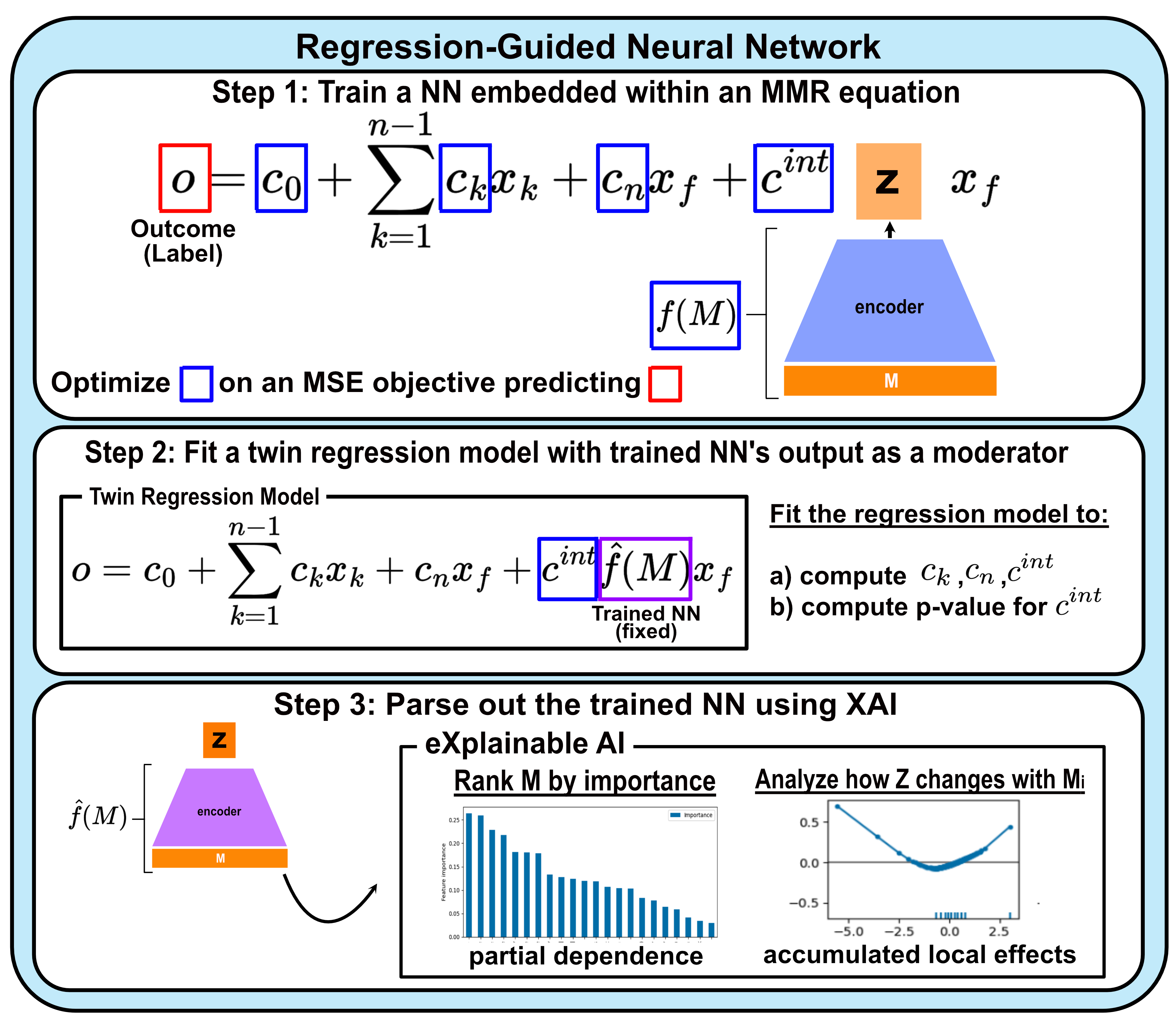}
    \caption{Overview of how a Regression-guided Neural Network (ReGNN) is trained and analyzed. First, a neural network is embedded within an MMR equation and trained to summarize the moderators (\textit{M}). Then, a twin regression model with the same equation used to train the network is fitted to compute its regression coefficients as well as their statistical significance. If a meaningful interaction is determined to be found, the trained neural network is parsed using explainable AI tools such as partial dependence.}
    \label{fig:regnn}
\end{figure}

In this paper, we propose a hybrid method that circumvents the drawbacks of both by training a neural network within the additive structure of MMR to summarize the predictors into a single (or a small number of) moderator that interacts with the focal predictor \textit{x$_{f}$}. Figure \ref{fig:regnn} provides an overview of how this hybrid structure embeds a neural network inside a regression model. Briefly, we set up a regression equation with linear terms and an interaction term just like an MMR, where the interaction term is constructed by multiplying the focal predictor \textit{x$_{f}$} with outputs of a neural network \textit{f(M)}. Here, we co-optimize the parameters of the neural network \textit{f} and the regression coefficients on a mean-squared-error objective using gradient descent. After the training is done, we use the same regression equation to create and fit a twin-regression model with the output of the trained neural network taken as the moderator for the focal predictor. Consequently, we are able to use all statistical tools available to analyze a regression model as usual. Finally, once we determine that a significant interaction between the summarized moderator and the focal predictor is found, we use partial-dependence-based importance rankings \cite{friedmanGreedyFunctionApproximation2001} and accumulated local effects \cite{apleyVisualizingEffectsPredictor2019a} to understand which predictors take larger influence on the trained neural network. 

Intuitively, the additive structure of the regression equation forces the neural network to approximate all higher-order effects through the interaction term, and thus producing a reduced-dimensional representation of the input predictors that moderate the effect of the focal predictor \textit{x$_{f}$}. Unlike other supervised algorithms, because this hybrid structure depends on the regression loss which acts only as proxy supervision to guide the neural network, we name this method a Regression-Guided Neural Network (ReGNN).

The following outlines the highlights of our contribution:
\begin{itemize}
  \item We propose a hybrid method combining regression and neural network, named Regression-Guided Neural Network (ReGNN), that uses regression loss as a surrogate supervision to arrive at a meaningful latent representation, specifically one that moderates the effect of a focal predictor \textit{x$_{f}$}. 
  \item We demonstrate that ReGNN is capable of uncovering population heterogeneity that would be hidden if traditional regression or machine learning models are utilized. We showcase this through asking important questions in public health context with an example dataset. 
\end{itemize}
\section{Related Works}
ReGNN, at its core, is a model that employs functional decomposition. \citet{hookerDiscoveringAdditiveStructure2004} describes a method discovering an additive structure for a black-box model. Such framework is referred to as functional ANOVA, and several following works propose to generalize this method \cite{huangFunctionalANOVAModels1998, hookerGeneralizedFunctionalANOVA2007}. While these techniques are developed as a post-hoc method to understand a trained black-box model, our proposed method brings function decomposition to its structure, making it inherently explainable.

On the other hand, there have been efforts to adapt an inherently transparent model to better capture the interactions among predictors. \citet{limLearningInteractionsHierarchical2015a} proposes hierarchical group-lasso regularization to sparsely find and hierarchically grow the number of interaction terms included within a regression model. ReGNN differs in that it focuses on a single predictor and learns to approximate all higher-ordered terms with other predictors that could potentially interact with it.

\citet{hornungInteractionForestsIdentifying2022a} proposes an interaction forests model that extends tree-based model by introducing splits that look at two variables simultaneously. While these methods are also inherently explainable and focus on discovering interaction effects, the tree model inherently does not decompose its findings into an additive structure, which makes it difficult to translate to the language of regression. 

\section{Methodology}
Moderated multiple regression is a simple generalization of linear regression that allows one to capture systematic differences in how a focal predictor affects an outcome. Suppose you are interested in understanding the relationship between an outcome variable \textit{o} and independent variables \textit{x$_{k}$}. Then, one would setup a linear equation and inspect its coefficients:
\begin{equation} \label{eq:lr}
    o = c_0 + \sum_{k=1}^{n} c_k x_k
\end{equation}
where \textit{c$_{k}$} are regression coefficients and \textit{c$_{0}$} is an intercept. Moderated multiple regression augments the above equation by adding interaction terms \textit{x$_{k}$x$_{f}$}: 

\begin{equation} \label{eq:mmr}
    o = c_{0} + \sum_{k=1}^{n-1} c_{k}x_{k} + c_{n}x_{f} + \sum_{k=1}^{n-1} c^{int}_{k}x_{k}x_{f}
\end{equation}
where \textit{x$_{f}$} is the focal predictor, \textit{c$_{n}$} is the regression coefficient for \textit{x$_{f}$}, and \textit{c$^{int}_{k}$} are regression coefficients for interaction terms. Without loss of generality, \textit{x$_{n}$} from the linear equation (\ref{eq:lr}) is assumed to be the focal predictor here. In practice, however, throwing in all remaining predictors into the interaction terms causes multicollinearity and reduction of the model's statistical power, making it difficult to find any meaningfully significant moderation effect.

Regression-Guided Neural Network (ReGNN) leverages neural networks to create a parsimonious regression model with just a single (or a few) interaction terms. Suppose we take the equation \ref{eq:mmr} and modify it by replacing all interaction terms with a single term that multiplies the focal predictor \textit{$x_{f}$} with an output of a neural network:
\begin{equation}\label{eq:regnn}
    o = c_0 + \sum_{k=1}^{n-1} c_{k}x_{k} + c_{n}x_{f}  + c^{int}f(M)x_{f}
\end{equation}
where \textit{f} represents the function approximated by the neural network, $c^{int}$ is a regression coefficient for the replaced interaction term, and \textit{M} is a vector representation of moderators \textit{$x_{1}, x_{2}, ..., x_{n-1}$}. Because the neural network considers the entire set of input variables to figure out all higher-order interactions with the focal predictor rather than each of them separately, it frees practitioners from having to guess which, how many, and which order of interaction terms to include in their model.

We outline the training and evaluation procedure for ReGNN in the following. We first set up the equation \ref{eq:regnn} with attached neural network in pytorch \cite{paszkePyTorchImperativeStyle2019a}, where the regression coefficients $c_{k}$ and \textit{$c^{int}$} and parameters of the neural network \textit{f} set as learnable parameters. We use gradient descent to optimize a mean-squared-error (MSE) objective, predicting an outcome \textit{o}. For datasets with weights for each sample, which is common in many survey studies, we use weighted-MSE as our objective.

At the end of each epoch, we set up a twin regression model on Stata \cite{statacorpStataStatisticalSoftware2023} with the same equation, with the output of the trained neural network taken as the only moderator for the focal predictor. We fit this twin regression model, evaluating the statistical significance of the regression coefficient for the interaction term (\textit{$c^{int}$}). While the network is trained on the MSE objective, our goal during training is to have the p-value for this regression coefficient to be as low as possible. Note that the p-values test for the statistical significance of the coefficients against a null hypothesis where the coefficients equal zero. The null hypothesis is rejected if the p-value falls under 0.05. Measures such as the twin regression model's r-squared and the variance inflation factors (VIF) are also observed. 

We run the regression model on a test dataset that is hidden from the neural network during training to evaluate its performance. This is to prevent an overfitting situation where the p-value for the interaction term's regression coefficient is low only on the train set while diverging on the test set. While calling this overfitting may be a bit awkward, as usually it refers to cases where the network's train and test scores on the training objective diverge, this is to prevent the network from just memorizing to predict the residual, or variances unexplained by the linear terms, from its inputs rather than coming up with a representation that is actually meaningful. 

Once the training is done, we run all samples through the trained neural network to produce a summary variable for each sample. The twin regression model is fit on all samples with the produced outputs taken as a moderator. We report the fitted regression coefficients as well as their confidence interval and p-value. If a statistically significant (p $<$ 0.05) interaction is found, our interest is now on how the neural network produced the summary variable.

With a significant moderation effect found, we finally use tools such as accumulated local effects (ALE) and partial dependence to understand which and how the input predictors \textit{$x_{1}, x_{2}, ..., x_{n-1}$} influence the trained neural network's output. 

\subsection{Advantages of using ReGNN over traditional regression or machine learning models}
In comparison to a traditional MMR, ReGNN is able to not only circumvent the multicollinearity problem by non-linearly combining the moderator variables, but also is capable of discovering hidden interaction effects. Consider a predictor \textit{$x_{l}$} that moderates the focal predictor not in an isolated single interaction term but instead dispersed into multiple higher-ordered terms in a true model:
\begin{equation} \label{eq:mmr_dispersed}
    o = c_{0} + \sum_{k=1}^{n-1} c_{k}x_{k} + c_{n}x_{f} + \sum_{k=1}^{m} c^{int}_{lk}x_{k}x_{l}x_{f}
\end{equation}
where m is some integer smaller than \textit{n-1}, and \textit{$c^{int}_{lk}$} are regression coefficients for three-way interaction terms \textit{$x_{k}x_{l}x_{f}$}. Suppose some \textit{$c^{int}_{lk}$} are both positively and negatively valued. If trying to estimate this using just a two-way interaction term \textit{$x_{l}x_{f}$}, its regression coefficient \textit{$c^{int}_{l}$} would be small, not sufficiently capturing the moderation effect of \textit{$x_l$}. Therefore, it becomes practically impossible to capture such hidden interaction effects using MMR without knowing the true model. On the other hand, because ReGNN takes the entire multivariate space to estimate and collect all moderation effects of each predictor across all terms, it is able to capture such hidden moderation effect.

This advantage becomes a disadvantage when traditional machine learning is used to predict the outcome directly. In ReGNN, already function decomposed structure of the equation makes it easy for the embedded neural network to isolate and approximate higher-order-terms only. Without this decomposed structure, it becomes hard to disentangle a model post hoc. For instance, partial dependence of a single variable includes not only its main effect, but also all of its influence on the higher-ordered-terms \cite{molnar2022}. Furthermore, Friedman's H-statistics can easily be inflated with correlated input predictors \cite{inglisVisualizingVariableImportance2021}.

\section{Experiment}
In this section, we showcase the use of ReGNN through applying it to a case study, uncovering population heterogeneity in the health effects of air pollution (PM2.5) on cognitive functioning scores. We demonstrate that ReGNN is capable of finding moderation effects that would otherwise be hidden using traditional methods by comparing its results to traditional MMR models.

\subsection{Dataset}
We compile four primary datasets for our analysis. First, we utilize individual-level data from the 2016 Health and Retirement Study (HRS), a nationally representative survey of U.S. adults aged 50 and older \cite{sonnegaCohortProfileHealth2014}. This dataset provides extensive sociodemographic and health status information, which we employ as health outcome and moderating variables. Second, we acquire 2016 daily Particulate Matter 2.5 (PM2.5) concentration data from the U.S. Environmental Protection Agency’s Fused Air Quality Surface Using Downscaling (FAQSD) files. The FAQSD datasets provide daily estimates for local concentrations of PM2.5 across the contiguous U.S. using a data fusion method, or the downscaler model. Third, we use neighborhood socioeconomic data from the U.S. Census Bureau’s 2014-2018 American Community Survey (ACS) 5-year estimates data. Both PM2.5 and ACS data, available at the census tract level for HRS respondents from the HRS-Contextual Data Resource \cite{ailshireUSDecennialCensus2020, ailshireParticulateMatterFAQSD2020}, are merged with the 2016 HRS sample records using their geocoded address-derived census tract information and interview dates. Finally, we obtain land cover data based on the U.S. Geological Survey National Land Cover Database, available from the National Neighborhood Data Archive at the census tract level \cite{melendezNationalNeighborhoodData2022}, and linked these data to the 2016 HRS sample.

\subsubsection{Health outcome: Cognitive function}
HRS respondents' cognitive functioning is assessed using the modified Telephone Interview for Cognitive Status (TICSm) \cite{ofstedalDocumentationCognitiveFunctioning2005}. The TICSm evaluates three cognitive domains: memory (test of 10 words immediate and delayed recall, range: 0-20 points), working memory (test of serial 7s subtraction, 0-5 points), and speed of mental processing (test of backward counting, 0-2 points). The composite summary score ranges from 0 to 27.  

\subsubsection{Focal predictor: Air pollution (PM2.5)}
For each HRS respondent, we calculate the mean concentration of residential PM\textsubscript{2.5 } ($\mu g/m^3$) at the census tract level over the past 30 days preceding their interview dates. 

\subsubsection{Other input variables}
Other predicting and moderating variables are selected a priori, based on their potential roles as effect modifiers in the association between air pollution and cognitive function \cite{delgado-saborit_critical_2021} as well as their roles as main risk (or protective) factors for the cognitive function. 
Individual-level predictors include age, gender (men and women), race/ethnicity (non-Hispanic White, non-Hispanic Black, Hispanic, and non-Hispanic Other), education levels (less than high school, high school, some college, and Bachelor’s and above), household income (log-transformed), household non-housing financial wealth (transformed using an inverse hyperbolic sine), BMI index, five self-reported chronic conditions diagnosed by a physician (i.e., stroke, heart disease, lung disease, diabetes, hypertension), depressive mood, light, moderate, and vigorous physical activity levels (never, $<$1 per month, 1-3 times per month, 1-2 times per week, and 3+ times per week). 

Area-level predictors include neighborhood socioeconomic status, urban/rural residence, and green space index. Neighborhood socioeconomic status at census tract level is assessed through factor analysis with orthogonal rotation on 10 indicators (i.e., \% without high school degree among the population aged 25+, \% of persons with below poverty level income, \% unemployed among the population aged 16+, \% households on public assistance income, \% families with children and in poverty, \% homeowners, \% with a bachelor’s degree among the population aged 25+, \% employed in management, professional, and related occupations, median house value, and median household income), resulting in the extraction of two factors reflecting neighborhood affluence and deprivation. Green space within each census tract is quantified by calculating the percentage of land cover classified as various types of vegetation and wetlands (e.g., deciduous forest, evergreen forest, mixed forest, shrub/scrub, herbaceous, and hay/pasture), with the index ranging from 0 to 100, where higher values indicate greater coverage of green space. 
\begin{figure}
    \centering
    \includegraphics[width=\linewidth]{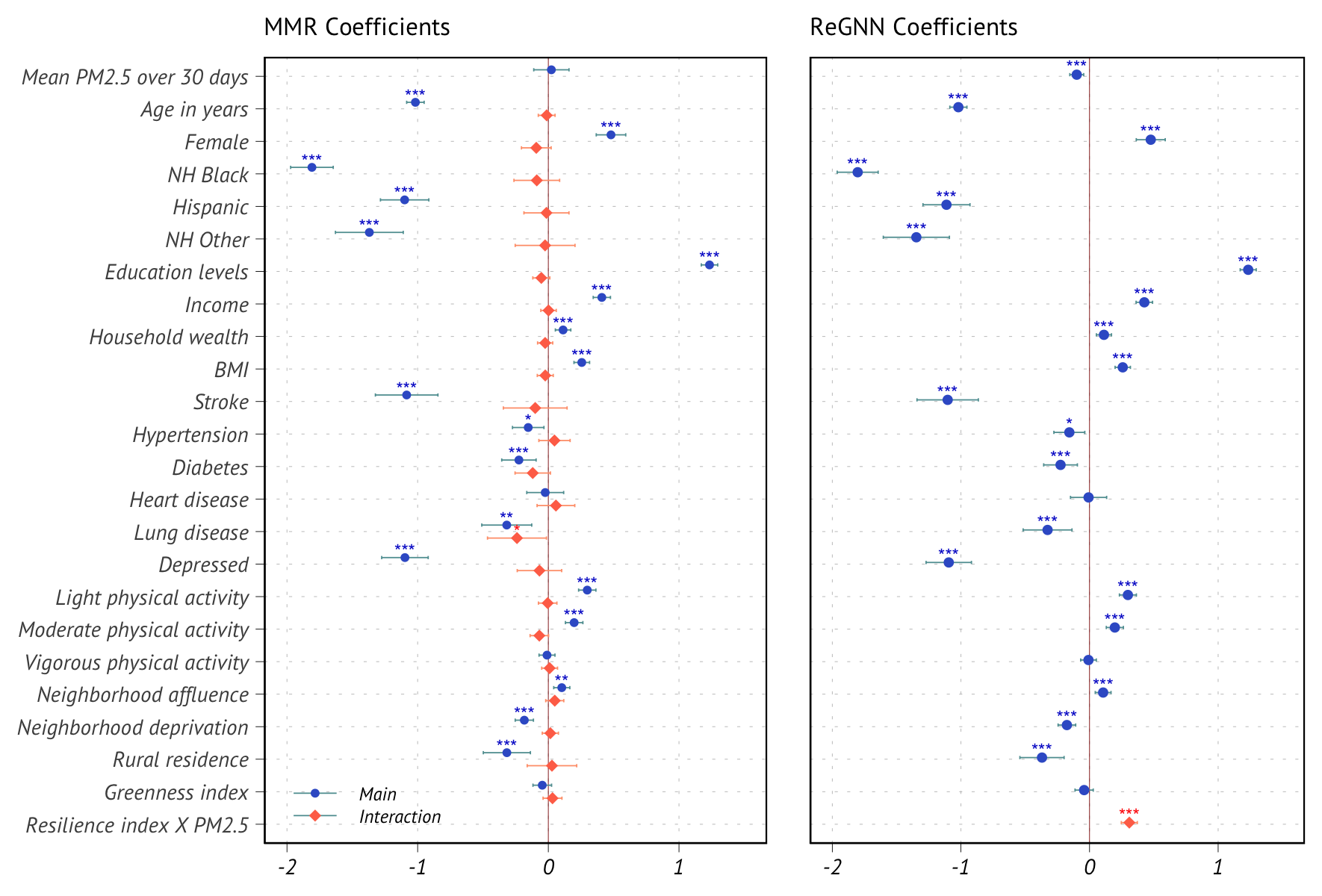}
    \caption{Regression coefficients comparing MMR models with (left) all predictors included as moderators ($r^2$ = 0.3135) and (right) only output of the trained neural network, which we name resilience index, included as moderator ($r^2$ = 0.3148). The position along the x-axis tells each coefficient's value, along with its confidence interval, which is indicated by the error bar. Significance levels are indicated with asterisks ($^{***}$: p $<$ 0.001; $^{**}$: p $<$ 0.01; $^{*}$: p$<$0.05)}
    \label{fig:regression_comparison}
\end{figure}
\subsection{Experimental setup}
An ensemble of five MLPs with layer sizes of \textit{($D_{in}$, 30, 10, 2, 1)}, where $D_{in}$ represents the number of moderators, is utilized. Each layer is followed by a GeLU activation function, except for the last layer. Outputs of each model in the ensemble are averaged and then batch-normalized. AdamW optimizer with learning rate of 0.005 is used. Weight decay of 0.01 is applied only to the regression coefficients. No regularization on the neural network parameters is applied in this model. Dropout rate of 0.2 is used during training. No dropout is applied during the evaluation phase at the end of each epoch. 

We randomly divide the dataset into training and testing sets, with a 70/30 split. We standardize continuous and ordinal variables, while converting categorical variables using one-hot encoding. We weight the mean-squared-error loss using survey weights, which compensate for different selection probabilities per sample. We train ReGNN for 150 epochs on the training set, while evaluating the model's performance on both the training and testing sets at the end of every epoch. 
\begin{figure}
    \centering
    \includegraphics[width=\linewidth]{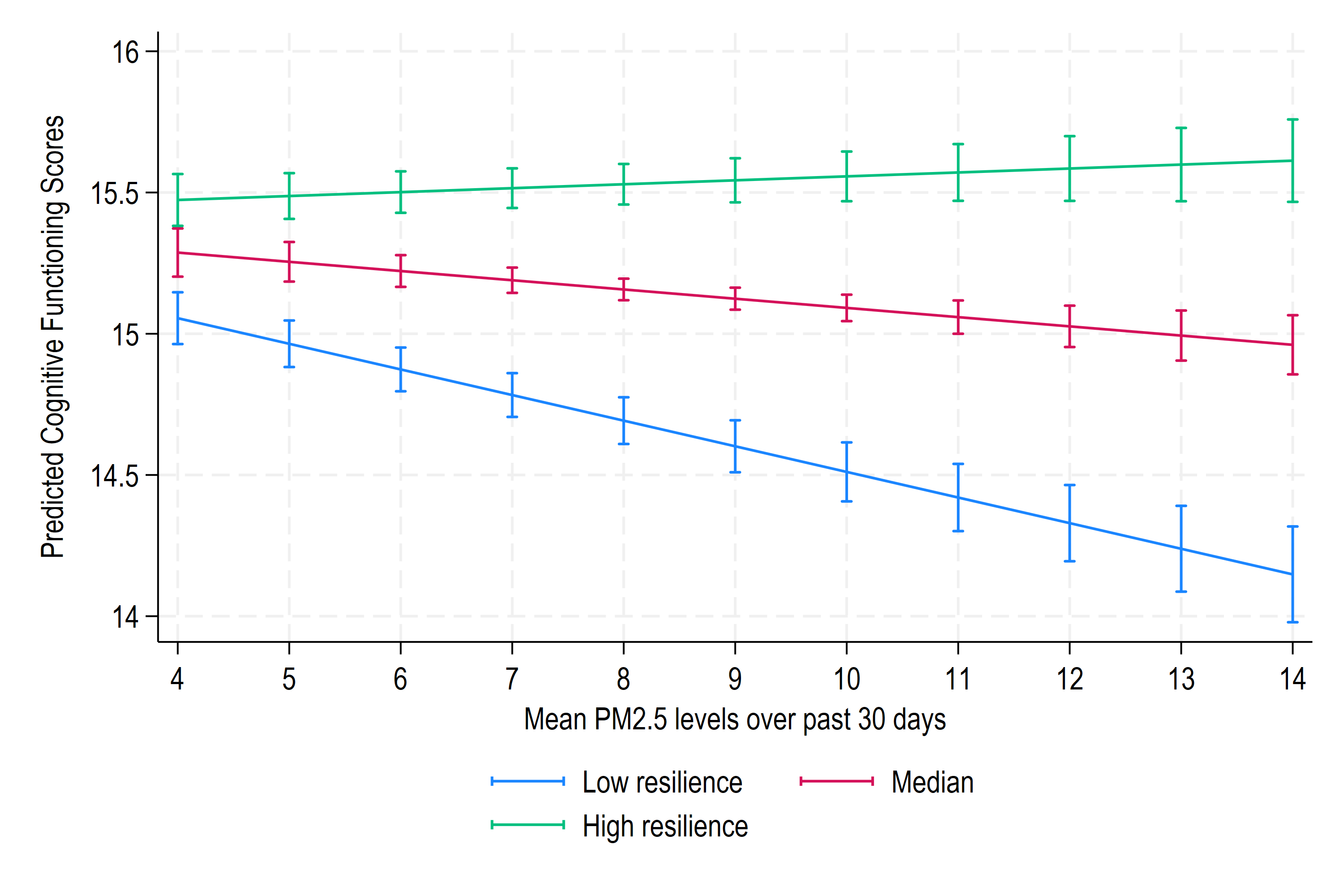}
    \caption{Predicted cognitive functioning scores based on MMR fitted with ReGNN-produced index (resilience index) as the moderating variable. Holding other independent variables to their means, the fitted MMR is used to predict the means and errors of the predicted cognitive scores for differing levels of PM2.5. Groups with low (bottom 10th percentile), median, and high resilience index (top 10th percentile) are separately plotted to show the moderating effect of the index on the effects of PM2.5 on the predicted cognitive score.}
    \label{fig:interaction}
\end{figure}
\subsection{Results}
We show that the output of the trained neural network significantly moderates the effect of air pollution on cognitive functioning scores. Figure \ref{fig:regression_comparison} displays the regression coefficients for MMR models, one fitted with all predictors included as moderators (left) and another with the outputs of the trained neural network, which we name the resilience index, included as the only moderating variable for PM2.5 (right). All of the predictors are added as independent variables in both models.

While most of the regression coefficients for the interaction terms in the MMR model are not significant (or only weakly significant for lung disease), we observe that ReGNN successfully produces a summary variable, which we name the resilience index, that moderates the effect of PM2.5 with high statistical significance (p $<$ 0.001). In addition, we report that the variance inflation factors (VIF) are higher for the traditional MMR model (maximum VIF = 6.48, with centering), in comparison to that fitted with resilience index (maximum VIF = 2.21). This shows that the twin-regression model is not only parsimonious, but also is free from the structural multicollinearity problem.

We experiment by including each predictor separately as the sole moderator in MMR to demonstrate that even with regression models with similar complexity, hidden interaction effects cannot be found with MMR. We discover only the same and no additional findings using single-interaction MMR models, where only the model with lung disease included has a significant interaction term (see supplementary materials).

The moderating effect of the resilience index is better displayed in figure \ref{fig:interaction}. Here, we plot the means and errors of the predicted cognitive functioning scores using the fitted twin regression model, for differing levels of PM2.5. Other independent variables are held at their means in getting the predicted cognitive scores. We observe that the deteriorating health effect of air pollution on cognition is significantly reduced for those with a higher resilience index (cyan, top 10th percentile) while it is inflated for those who score low on the resilience index (blue, bottom 10th percentile). Because the error bars for each predicted point do not overlap across the three groups, we conclude that the resilience index significantly moderates the effect of PM2.5 on cognition.

\subsubsection{Subtle changes that occur after the losses plateau drive the representation learning in ReGNN}
\begin{figure}
    \centering
    \includegraphics[width=\linewidth]{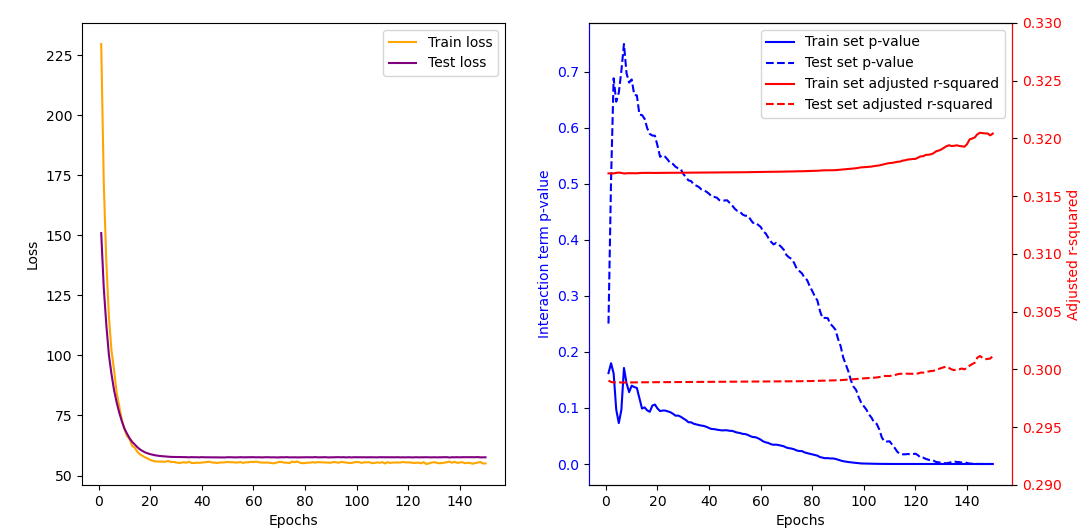}
    \caption{Trajectories of model performance metrics during the training session. (Left) Train and test losses; (Right) the p-values of the interaction term's coefficient (blue) and adjusted R-squared (red) of the twin-MMR model, fitted to the train set (solid line) and test set (dotted line) respectively. While the losses plateaus after 20 epochs, the p-values significantly decrease afterward.}
    \label{fig:trajectory}
\end{figure}
Here, we illustrate how ReGNN arrives at a representation that we want with the guidance of its objective function. Figure \ref{fig:trajectory} plots the train and test losses (left), the p-values of the interaction term's coefficient, and the adjusted R-squared values for the twin regression models fit to train and test datasets (right). We observe that while the train and test losses plateau after 20 epochs, the p-values of the interaction term's coefficient decay significantly afterward. 

\begin{figure}
    \centering
    \includegraphics[width=\linewidth]{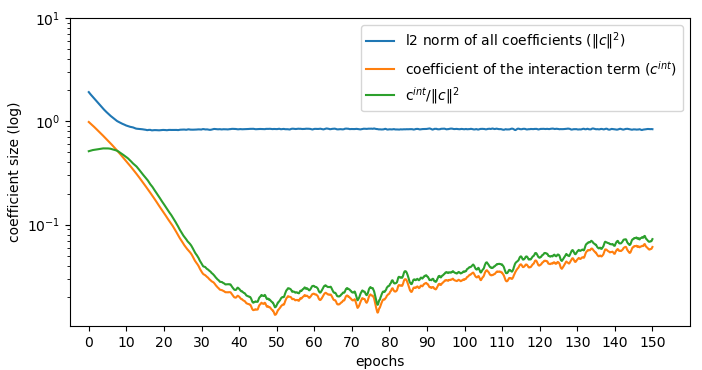}
    \caption{Trajectories of log-magnitudes of ReGNN's regression coefficients. Blue shows the L2 norm of all regression coefficients ($c_{k}$, $c_{n}$, and $c^{int}$), orange shows the magnitude of the coefficient for the interaction term ($c^{int}$), and the green shows the ratio of the two. While the l2 norm stops decaying early on, $c^{int}$ decays until it reaches a minimum, and bounces back up. The ratio shows that the overall magnitude (denominator) stays almost the same after 20 epochs.}
    \label{fig:coefficients}
\end{figure}

Why is the ReGNN's representation learning decoupled from the decay of its objective function? Figure  \ref{fig:coefficients} captures how ReGNN changes after the initial decay of the loss function. Here, we plot the trajectories of the l2 norm of all regression coefficients ($c_{k}$, $c_{n}$, and $c^{int}$) (blue), and the regression coefficient for the interaction term only (orange), and their ratios (green). The trajectories show us that the parameters subtly change within ReGNN even after the losses plateau. ReGNN seems to go through different learning phases. During the initial 20 epochs, ReGNN adjusts the overall magnitude of its regression coefficients ($c_{k}$, $c_{n}$, and $c^{int}$) to achieve a near-minimum in MSE. This corresponds to the initial sharp drop in MSE and the l2 norm of coefficients shown in blue. As the training proceeds, we observe that \textit{$c^{int}$} decreases and bounces back up (orange), while the l2 norm of all coefficients stay the same. We hypothesize that when the neural network is producing noisy output, ReGNN tries to minimize MSE by suppressing its effect through a decay in \textit{$c^{int}$}. Then, once the network starts to produce output that explains the residual variance through the interaction term, the proportion explained by the interaction grows, corresponding to increase in \textit{$c^{int}$} and the model's r-squared (red, figure \ref{fig:trajectory}), and the decrease in p-value for the interaction term. This shows that subtle changes that occur after the decay in its objective drive the representation learning in ReGNN.

\subsubsection{Explainable AI techniques help understand how each predictor contributes to the generation of the summary index }
With regression coefficients showing a significant moderation effect of the resilience index, it remains now to understand how ReGNN produces them. Here, we employ publicly available explainable AI tools to parse which predictors contributed to the network's assignment of values for this index. 

\begin{figure}
    \centering
    \includegraphics[width=\linewidth]{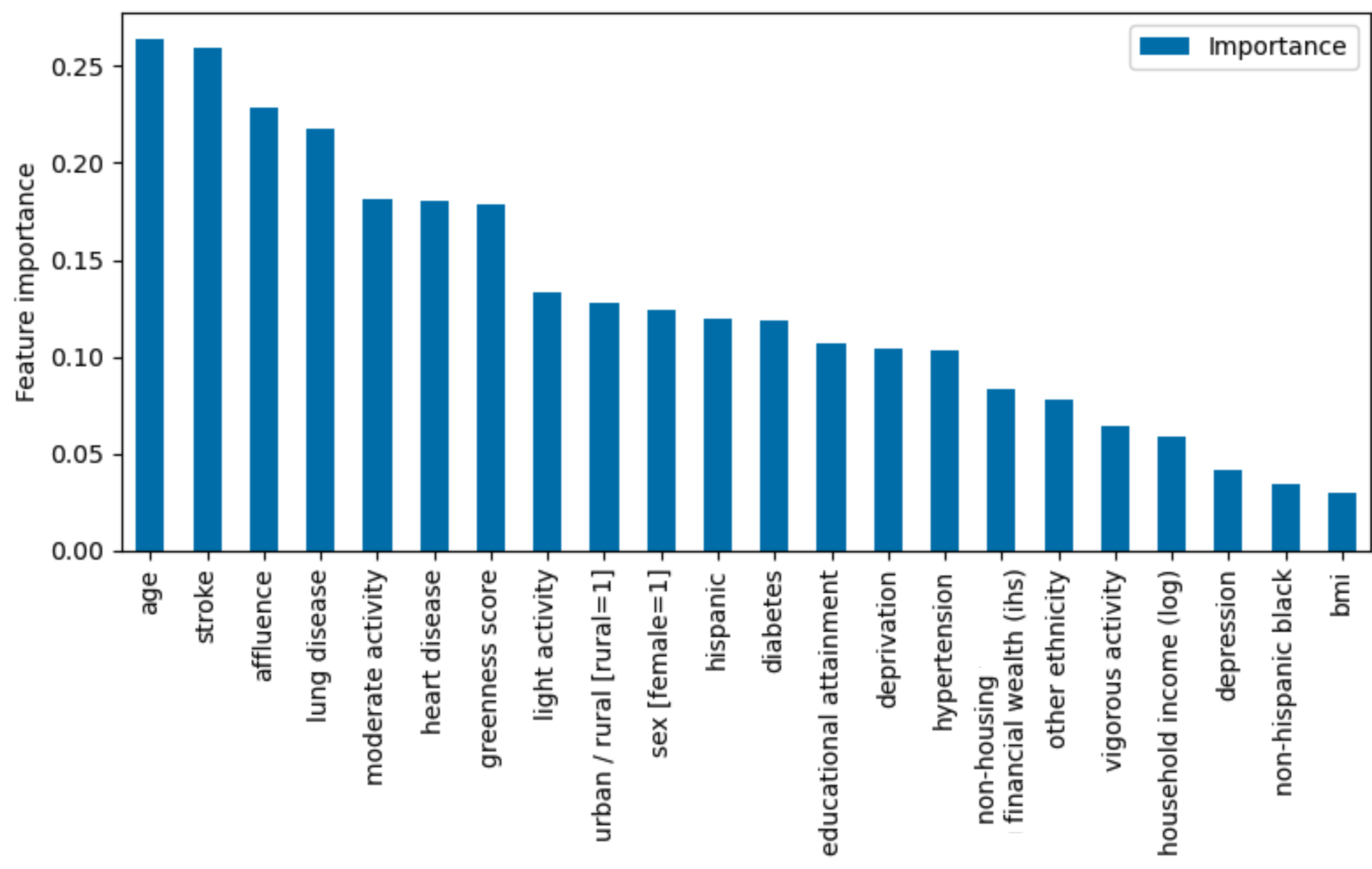}
    \caption{Partial-dependence-based feature importance. 1000 random samples are selected to estimate the partial dependence. The height displays how much the trained neural network's output (resilience index, std $\approx$ 1) changes when each feature is varied.}
    \label{fig:partial_dependence}
\end{figure}

Figure \ref{fig:partial_dependence} rank orders the input predictors based on the partial dependence \cite{friedmanGreedyFunctionApproximation2001}. This approximates an averaged global behavior of the trained model and orders the input variables based on how much the output changes on average. Other tools such as Shapley additive explanations (SHAP) can be used to create a similar display \cite{lundbergUnifiedApproachInterpreting2017}. 

We report that the order based on the feature importance can change. This is attributable to randomness introduced while doing a train / test split, training the neural network, and estimating the importance measure from randomly selected samples with Monte Carlo. While not in the scope of this paper, we can adopt methods such as permutation tests \cite{miPermutationbasedIdentificationImportant2021} to decide a cut-off, testing for which feature importance measures are high enough to rule out a null hypothesis, tested against a distribution of the chance-level importance scores with each feature randomly permuted. 

To better understand in how the produced index gets influenced by each feature, we use accumulated local effects (ALE) plot to display how a local change in a feature leads to an averaged change in the output \cite{apleyVisualizingEffectsPredictor2019a}. Note that ALE corrects for the distortion that can result from correlated input features in partial dependence plots \cite{molnar2022}. Figure \ref{fig:ale} displays ALE plots for a select set of features based on their feature importance. (See supplementary materials for the exhaustive set.) The small vertical lines along the x axis show the distribution of the sampled feature. 

With ALE, it is possible to not only observe the direction of change in the output, but also qualitatively judge the magnitude of the influence for ranges of values that most commonly occur in the dataset. For example, while affluence seems to largely affect the model output at a first glance, we observe how the magnitude of the influence is reduced for the range where most samples are distributed (-1 to 1). While these two do not exhaust the list of explainable AI tools we could apply, they provide enough information to qualitatively judge what is going on within the model.

We conduct a review of relevant literature to assess how aligned the high-ranked features are with the past findings. We find that the directions of the influences for many features are aligned: less affluent neighborhood socioeconomic status exacerbates the effects of air pollution on cognitive function \cite{li_neighborhood_2022} and decline \cite{christensen_complex_2022}; increased levels of physical activity intensify the adverse effects of air pollutants on cognitive function \cite{li_interaction_2024, liu_joint_2023}; high residential greenness areas attenuate the associations between high NO2 and poorer executive function \cite{wang_greenness_2023}; women are at a higher risk for decreased cognitive function associated with increased exposure to PM10 and PM2.5-10 \cite{kim_gender_2019}; Hispanics show less rapid pollution-associated decline in global cognitive function than non-Hispanic Whites \cite{kulick_long-term_2020}. However, while we identify several medical conditions as important contributors to the resilience index, few studies have empirically tested these as effect modifiers, and the findings are heterogeneous \cite{delgado-saborit_critical_2021}, complicating direct comparisons.
\begin{figure}
    \centering
    \includegraphics[width=\linewidth]{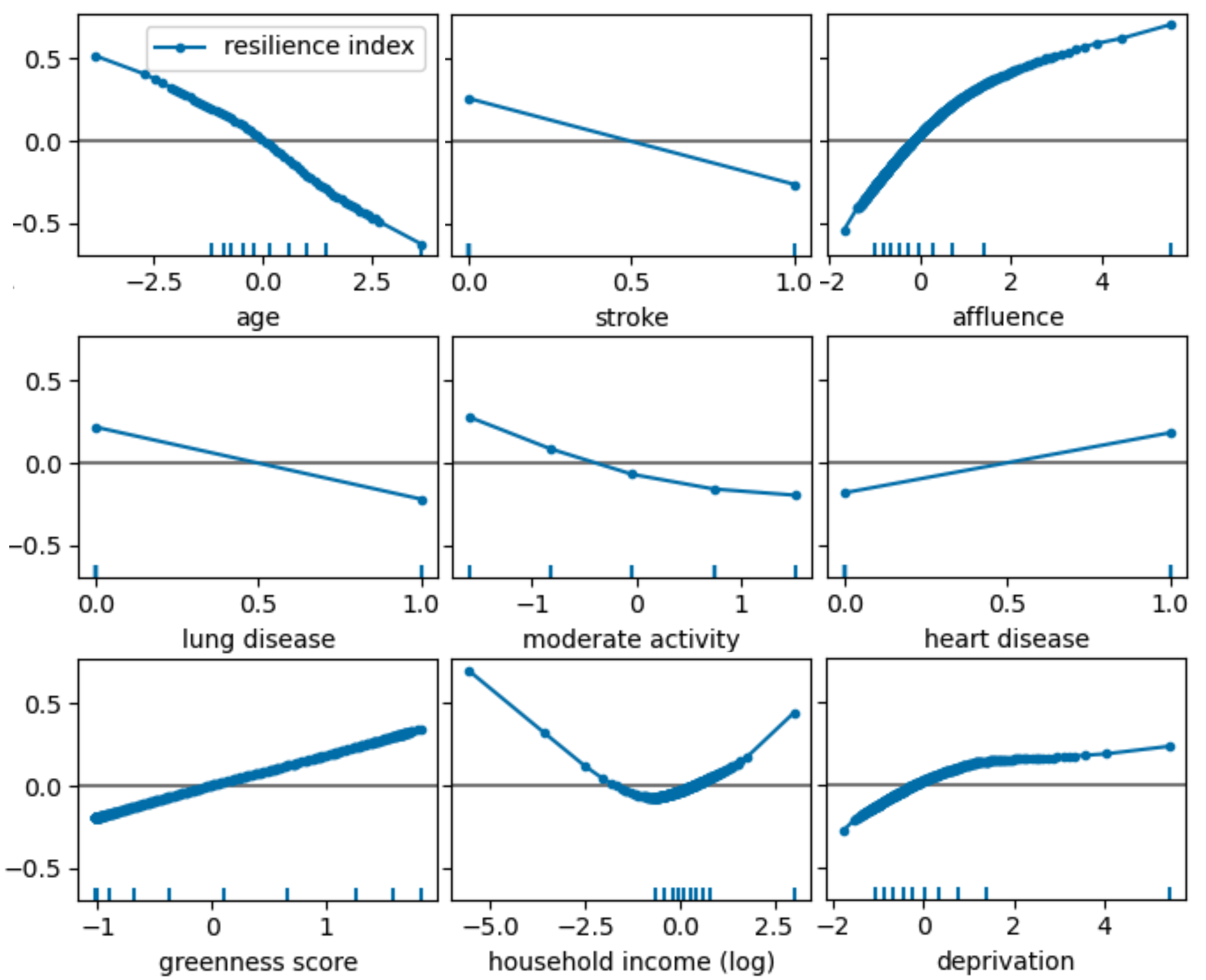}
    \caption{Accumulated local effects (ALE) for selected predictors. Vertical axes are in the units of the resilience index (std $\approx$1). Small vertical line segments along the horizontal axes show how the data for each feature are distributed.}
    \label{fig:ale}
\end{figure}

\section{Discussion and Future Works}
As environmental hazards grow more intense, prolonged, and severe, identifying and understanding factors driving heterogeneity in health risks posed by those hazards will become crucial. In this sense, ReGNN provides a transformative utility in that it is capable of capturing heterogeneity hidden within a complex multivariate space that no other traditional methods can uncover. 

In addition, because ReGNN quantifies susceptibility to health risks posed by an environmental hazard for each individual, it can potentially be used to come up with strategies that mitigate each individual's vulnerability to the hazard. While we present the simplest form of ReGNN in this paper, its concept that embeds a machine learning model within the traditional regression model can be expanded further to accommodate such potential. We seek to explore embedding other forms of encoders that can estimate the uncertainties along with the predictions (i.e. variational autoencoder) in the future.

Limitations in tools used to understand the trained encoder directly translates to the limitations of our method, however. While measures such as feature importance provide some qualitative picture of how the trained model works, it often is not sufficient for scientists seeking to understand the actual magnitudes of these measures with statistical robustness. We thus need to further investigate strategies that could facilitate robust scientific discoveries when parsing out the trained model.

\section{Acknowledgement}

\bibliography{regnn}

\end{document}